\documentclass[letterpaper, 10 pt, conference]{ieeeconf}  %

\IEEEoverridecommandlockouts                              %

\usepackage{graphicx} %

\usepackage{tikz}       %
\usepackage{lipsum}     %

\usepackage{caption}
\usepackage{cuted}   

\usepackage{graphicx}      %
\usepackage{amsmath,amssymb}
\usepackage{tabularx}
\usepackage{booktabs}      %
\usepackage{siunitx}       %
\usepackage{multirow}
\usepackage{makecell}
\usepackage{balance} %
\PassOptionsToPackage{hyphens}{url} %
\usepackage[bookmarks=true,colorlinks=true,citecolor=blue,linkcolor=blue]{hyperref}

\usepackage{threeparttable}

\usepackage{xspace}
\newif\ifshowcomments
\showcommentstrue
\showcommentsfalse  %
\ifshowcomments
    \usepackage[textsize=scriptsize]{todonotes}
    \newcommand{\fix}[1]{{\color{red} #1}}
\else
    \usepackage[disable,textsize=scriptsize]{todonotes}
    \newcommand{\fix}[1]{}
\fi

\newcommand{\todomw}[1]{\todo[fancyline,color=green!40]{MW: #1}\xspace}

\newcommand{\todovt}[1]{\todo[fancyline,color=cyan!40]{VT: #1}\xspace}

\newcommand{\alg}{\textsc{HapCompass}\xspace}%

\makeatletter
\let\NAT@parse\undefined
\makeatother
\usepackage[square, numbers, sort]{natbib}

\title{\LARGE \bf
\alg: A Rotational Haptic Device\\ for Contact-Rich Robotic Teleoperation 
}

\author{%
Xiangshan Tan$^{1}$, Jingtian Ji$^{1}$, Tianchong Jiang$^{1}$, Pedro Lopes$^{2}$, and Matthew R. Walter$^{1}$%
\thanks{$^{1}$Toyota Technological Institute at Chicago (TTIC), %
{\tt\small \{vincenttann, jijingtian, tianchongj, mwalter\}@ttic.edu}}%
\thanks{$^{2}$University of Chicago, %
{\tt\small pedrolopes@uchicago.edu}}%
\thanks{Accepted to IEEE International Conference on Robotics and Automation (ICRA), 2026.}%
}
\begin{document}
\maketitle
\thispagestyle{empty}
\pagestyle{empty}

\begin{strip}
    \centering
    \vspace{-5.7em} %
    
    \resizebox{\textwidth}{!}{\includegraphics{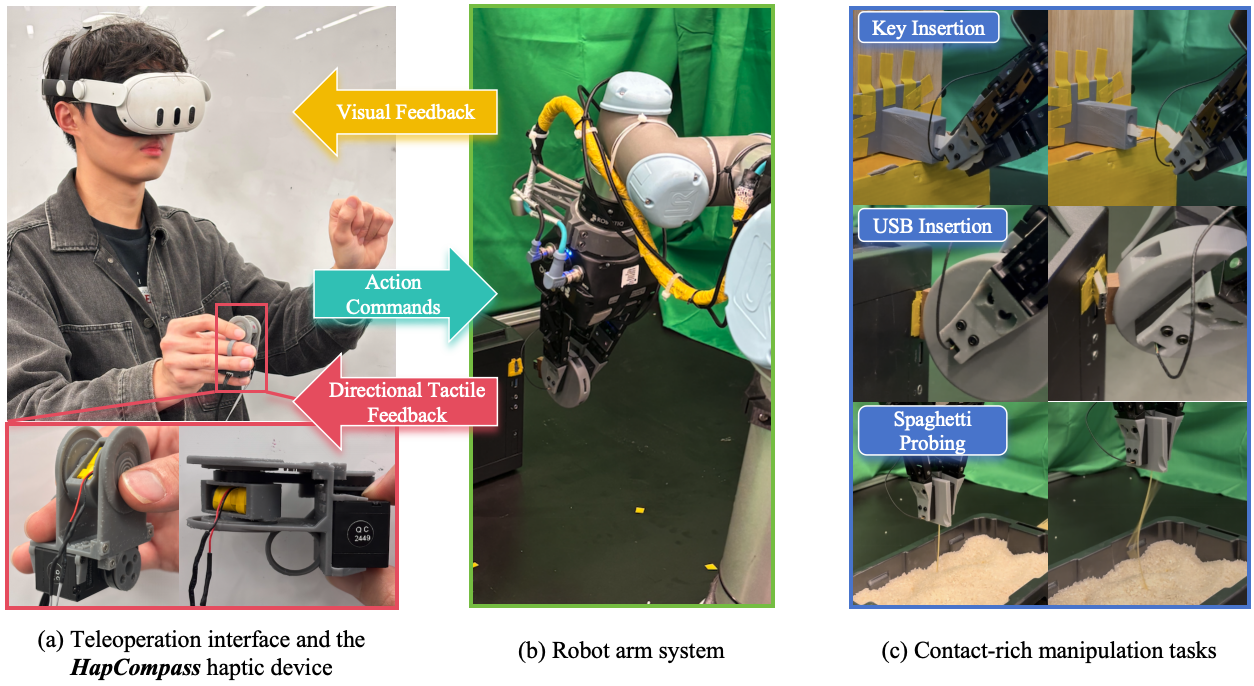}}
    \captionof{figure}{Overview of the \alg teleoperation system. (a) An operator controls the robot via hand tracking while wearing our novel \alg device, which renders directional haptic feedback from the robot's sensors. (b) The robot arm executes the manipulation commands. (c) We evaluate the system on three contact-rich tasks: Key Insertion, USB Insertion, and Spaghetti Probing, with (left) our method achieving higher success than (right) baseline methods.}
    \label{fig:teaser}
    \vspace{-0.5em} %
    
\end{strip}

\begin{abstract}
The contact-rich nature of manipulation makes it a significant challenge for robotic teleoperation. 
While haptic feedback is critical for contact-rich tasks, providing intuitive directional cues within wearable teleoperation interfaces remains a bottleneck.
Existing solutions, such as non-directional vibrations from handheld controllers, provide limited information, while vibrotactile arrays are prone to perceptual interference. %
To address these limitations, we propose \alg, a novel, low-cost wearable haptic device that renders 2D directional cues by mechanically rotating a single linear resonant actuator (LRA). %
We evaluated \alg's ability to convey directional cues to human operators and showed that it increased the success rate, decreased the completion time and the maximum contact force for teleoperated manipulation tasks when compared to vision-only and non-directional feedback baselines. 
Furthermore, we conducted a preliminary imitation-learning evaluation, suggesting that the directional feedback provided by \alg enhances the quality of demonstration data and, in turn, the trained policy. 
We release the design of the \alg device along with the code that implements our teleoperation interface: \url{https://ripl.github.io/HapCompass/}.

\end{abstract}

\section{Introduction}

Successfully executing contact-rich manipulation tasks, such as tight-clearance insertions (e.g., USB and Ethernet insertion~\cite{sliwowski2025reassemble}) or assembly in environments with little-to-no visibility (e.g., manipulating underwater connectors~\cite{baggeroer2018ocean})\todovt{I removed howe16 to save space}, is a key capability for intelligent robotic systems. However, enabling robots to perform these tasks,
either through direct teleoperation or by learning from human demonstrations (i.e., imitation learning), remains an open problem. 

A key aspect of this challenge lies in the inherent \textit{sensory asymmetry} between the human operator and the robot---while robots can be equipped with advanced tactile and force sensors~\cite{yuan2017gelsight,bhirangi2024anyskin}, operators sometimes lack corresponding channels to perceive these contact interactions.

In the established field of bilateral teleoperation~\cite{dale1993stability,lee2006passive,tavakoli2007high}, this asymmetry is addressed through control schemes that provide force feedback to the operator, typically using a second robotic arm or grounded haptic devices. While effective for providing directional force cues, these systems often require specialized, non-portable hardware that constrains the operator's workspace. 

In recent work, particularly within the robot learning community, there has been an increasing use of unilateral teleoperation interfaces for data collection, such as vision-based hand tracking and wearable VR controllers~\cite{qin2023anyteleop, cheng2024open,patil2025generalized}. These methods offer greater flexibility for teleoperation and data collection. However, due to the lack of informative haptic feedback, they often leave the human operators insensitive to contact feedback---forcing them into a low-bandwidth, vision-dominant decision loop---a major limitation given that human dexterity critically depends upon touch: classic motor control studies show that without cutaneous sensation, even simple tasks like lifting an object become clumsy and inefficient~\cite{johansson1984roles,westling1987responses}. 
Consequently, in many current practices, tactile signals are recorded only as a passive byproduct of vision-driven actions~\cite{pattabiraman2024learning}. This potentially weakens the correspondence between contact sensory events and operator intent, limiting the value of the data for learning.

While incorporating haptic feedback to these portable interfaces can improve both teleoperation performance and the quality of data for imitation learning~\cite{cuan2024leveraging, motamedi2016haptic}, signaling contact alone is insufficient for complex tasks. For contact-rich manipulation, feedback must convey not only that contact occurred but also the \emph{directional forces} involved, which is essential for precise adjustment during teleoperation.

Widely used non-directional vibrations from virtual reality (VR) controllers provide a low-dimensional signal by mapping contact force magnitude to vibration intensity~\cite{cuan2024leveraging,kamijo2024learning}. While this conveys the presence and intensity of a contact, it \textit{omits directional information}, which is often essential for guiding fine-grained adjustments.
Other common interfaces, such as haptic gloves~\cite{zhang2025doglove, li2024haptic}, typically render a single degree-of-freedom resistive force to each finger to simulate grasping. While useful for conveying contact via normal forces, this approach cannot communicate the direction of tangential forces. Alternatively, devices that use arrays of vibrotactile actuators to render directional cues~\cite{culbertson2017waves} can suffer from perceptual interference, where simultaneous vibrations mask each other and create ambiguous sensations.

To address these limitations, we propose \alg (Fig.~\ref{fig:teaser}), a novel, low-cost wearable haptic device designed to deliver 2D directional feedback. The core innovation of \alg is its principle of \textbf{mechanically rotating a single linear resonant actuator (LRA)}. The LRA is actuated with an asymmetric waveform, which creates a perceived pulling sensation along its actuation axis. By orienting this axis using a servo motor, \alg can render %
directional haptic cues in the $x$-$y$ plane.

This paper presents the design, integration, and evaluation of the \alg system. Our key contributions are:

\begin{itemize}
    \item A novel haptic device, \alg, that rotates a single LRA to render 2D directional cues.

    \item A teleoperation system that integrates visual and directional haptic feedback using \alg for contact-rich manipulation.
    
    \item A quantitative study of \alg's ability to convey directional haptic information to users.
    \item A teleoperation study showing that \alg enables users to perform contact-rich tasks more successfully and safely than with traditional interfaces.%

    \item A preliminary evaluation showing that demonstrations collected with \alg can improve the success of downstream imitation learning policies, while reducing peak contact forces and torques.

\end{itemize}

\section{Related Work}
\subsection{Directional Haptic Devices}
Prior work on fingertip-based systems that convey directional haptic cues can be broadly classified into two primary approaches~\cite{martin2024tactile}: those leveraging asymmetric vibrotactile stimulation and those that mechanically induce skin-stretch.

\textbf{Asymmetric Vibration} devices operate by generating an illusory ``pulling" sensation. This is achieved by driving a vibrotactile actuator with a signal that produces a higher peak acceleration in one direction than in the opposite, resulting in a perceivable, directional force cue~\cite{amemiya2005virtual, amemiya2008asymmetric, rekimoto2014traxion}. A common approach is to arrange multiple actuators in fixed, orthogonal orientations, allowing the system to render directional cues along the cardinal axes. For instance, WAVES~\cite{culbertson2017waves} uses three orthogonal Linear Resonant Actuators (LRAs) to present 3D translational cues. However, a significant limitation of such multi-actuator systems is the challenge of rendering arbitrary, off-axis directions. Attempting to synthesize a directional vector by simultaneously activating multiple actuators can lead to ambiguous perceptual cues or interference, muddling the intended direction. Other devices have focused on rendering force in a single, task-relevant direction, such as gravity, but are consequently limited to one degree of freedom~\cite{choi2017grabity, tanaka2020dualvib}. HapCube~\cite{kim2018hapcube} takes a more sophisticated approach and mechanically synthesizes 2D planar vibrations by coupling the motion of two orthogonal voice-coil actuators. While effective, this design requires precise synchronization of the driving signals in both phase and amplitude to ensure linear motion and avoid unintended elliptical vibrations.

\textbf{Skin-Stretch} devices aim to mimic the natural deformation of the fingerpad that occurs when contact forces are applied during object interaction. This can be accomplished through various mechanisms that move tangentially across the skin—known as ``tactors"—including moving platforms, belts, or actuated pins. For example, some designs use a single tactor capable of planar~\cite{girard2016haptip} or even three degree-of-freedom motion~\cite{leonardis20163} to render contact forces. While capable of producing strong and intuitive directional cues, skin stretch devices typically entail greater mechanical complexity and larger form factors compared to their vibrotactile counterparts~\cite{martin2024tactile}. Furthermore, their effectiveness can be subject to significant inter-user variability due to differences in individual skin properties, which complicates the mapping from device displacement to perceived force.

\subsection{Haptic Feedback in Robot Teleoperation}
While vision-only teleoperation systems~\cite{qin2023anyteleop, cheng2024open} have proven effective for a variety of tasks, existing work has shown that 
conveying the robot's sense of touch to the operator can substantially improve task performance, e.g., by reducing excessive contact forces and enabling faster completion times in manipulation tasks~\cite{motamedi2016haptic, kamijo2024learning,cuan2024leveraging}. The utility of such feedback, however, depends on the dimensionality and form of the rendered information.

Existing portable interfaces often provide limited directional cues.
The most common vibrotactile alerts, such as those found in commercial VR controllers, typically function as event cues---indicating the presence and approximate magnitude of contact---while conveying little about its spatial characteristics. A more advanced class of devices includes haptic gloves, which primarily render grasp-related normal forces~\cite{zhang2025doglove, li2024haptic}. By providing 1-DoF resistive force to each fingertip, they effectively simulate the sensation of holding a solid object. While useful for grasp stabilization, this feedback modality is incapable of communicating the direction of tangential forces, which plays an important role in contact-rich tasks such as sliding along a surface or performing tight-clearance insertions.

In the well-established field of bilateral teleoperation, directional force feedback is achieved through control architectures that ensure stability and transparency during teleoperation~\cite{dale1993stability,lee2006passive,tavakoli2007high}. These systems utilize a broad range of leader devices, from secondary robot arms~\cite{lenz2025nimbro} to specialized grounded haptic interfaces~\cite{poignonec2021simultaneous}. While these high-fidelity solutions provide multi-DoF force feedback, their high cost, limited workspace, and non-portable nature restrict their use.

\section{The HapCompass System}
\label{sec:system}

\alg provides a complete 
teleoperation system for contact-rich robotic manipulation comprising three core components: a wearable haptic device, an architecture for control and feedback flow, and a rendering algorithm that translates robot sensor data into directional haptic cues.

\subsection{Hardware Design}
\label{subsec:hardware}

\begin{figure}[!t]
    \centering
    \includegraphics[width=\columnwidth]{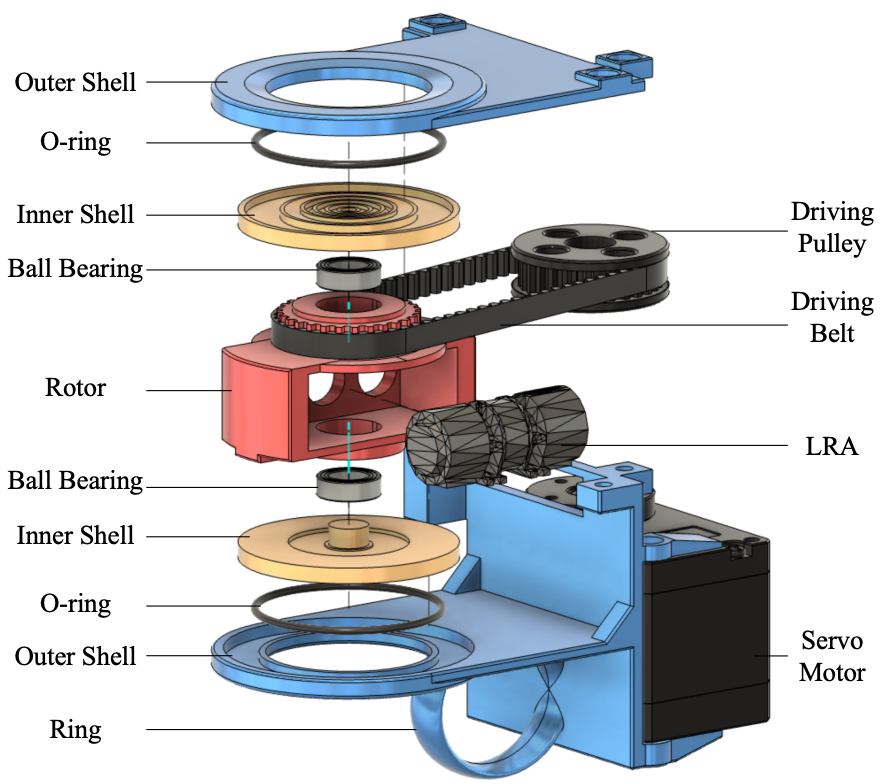}
    \caption{Exploded view of the \alg haptic device.}
    \label{fig:hardware}
\end{figure}
The \alg device (Fig.~\ref{fig:hardware}) is a novel wearable haptic interface. It attaches to the operator's hand via a ring on their middle finger, allowing the index finger and thumb to naturally rest on the upper and lower contact surfaces closest to its vibration actuator. Its key innovation lies in mechanically rotating a single linear resonant actuator (LRA) (\textit{Drake LFi} model, from \textit{TITAN}~\cite{titan-drake}, 
with up to 19G of impulse acceleration and 13.4G of steady-state acceleration) to render 2D directional haptic cues. 
The LRA is actuated with an asymmetric waveform, producing a perceived pulling sensation along its actuation axis.

The LRA is mounted within a 3D-printed rotor. This rotor is actuated by a compact servo motor (\textit{XL330}, from \textit{Dynamixel}~\cite{dynamixel-XL330}) via a driving belt and pulley transmission. 
To ensure low-friction rotation, the rotor is supported by two miniature ball bearings. The entire rotational assembly and the servo motor are housed within a central shell.

A key design challenge is to efficiently transmit vibrational energy to the user's fingertips while isolating it from the device's main structure. To achieve this, we designed a decoupled two-part housing that consists of an inner and an outer shell. As detailed in Fig.~\ref{fig:hardware}, the two shells are connected by an interlocking L-shaped structure that provides support along the $z$-axis. A gap between these interlocking features allows for small relative movements. O-rings seated within this gap act as compliant dampers. This combined mechanism helps to isolate the vibration from the main body held by the operator, reducing noise in the haptic feedback. %
The total cost to fabricate the device is approximately \$140 USD.

\subsection{Teleoperation System Architecture}
\label{subsec:architecture}

\begin{figure*}[!t]
    \centering
    \includegraphics[width=\textwidth]{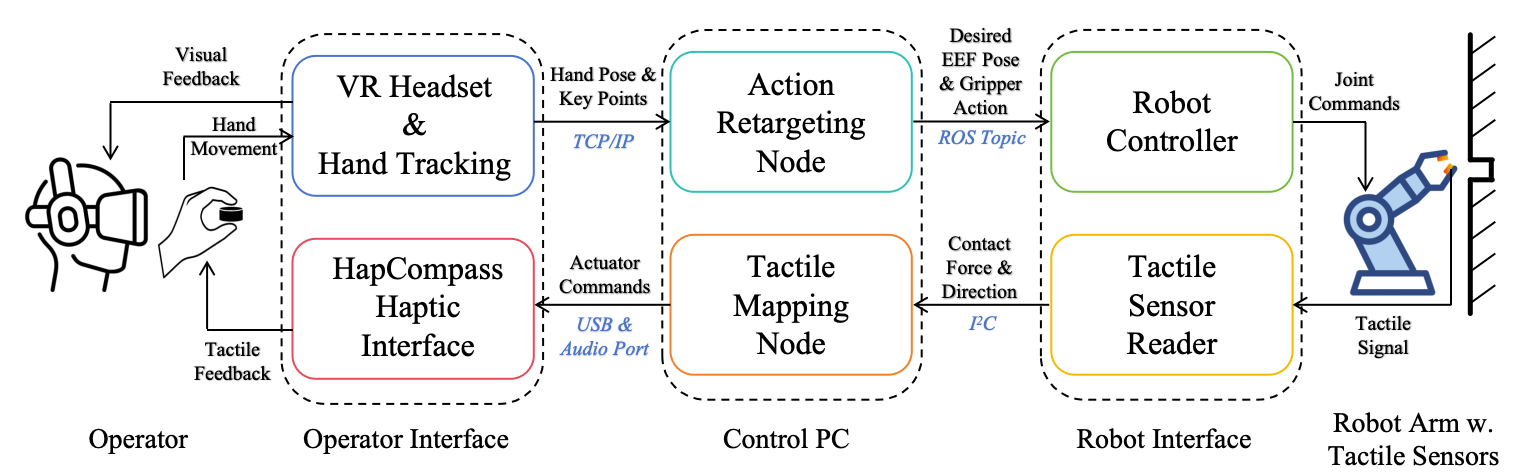}
    \caption{The overall \alg system architecture.}
    \label{fig:architecture}
\end{figure*}

The teleoperation system consists of a command path and a sensory feedback path, as shown in Fig.~\ref{fig:architecture}.
The command path begins with a Meta Quest 3 headset that tracks the operator's hands. Captured hand poses are streamed to a control PC, where an \textit{Action Retargeting Node} translates human hand motions into desired robot end-effector poses and gripper actions for execution.

The feedback path provides the operator with both visual and haptic information. Visual feedback is delivered through the Meta Quest 3's video-passthrough function, supplemented by a 2D video stream from a wrist-mounted camera on the robot for close-up views. 
Haptic feedback is obtained from the robot's three-fingered gripper, where each finger is integrated with a tactile sensor (\textit{TS-E-A}, from \textit{Tashan}~\cite{tashan-tsea}) 
that measures contact force and its direction. The robot arm is also equipped with a wrist-mounted force-torque (F/T) sensor  (\textit{FT-300}, from \textit{Robotiq}~\cite{robotiq-ft300})  that measures the net six-axis wrench exerted at the end-effector.%

When the robot interacts with the environment, the resulting tactile measurements are sent to a \textit{Tactile Mapping Node} on the control PC. This node processes the data and computes the appropriate commands for the \alg device, based on the mapping algorithm detailed in Section~\ref{subsec:rendering}.

\subsection{Haptic Rendering and Control}
\label{subsec:rendering}
The haptic rendering pipeline translates the measured 3D contact forces into 2D directional cues on the \alg device. 
The system primarily utilizes the force readings from the fingertip tactile sensors.

To ensure that the haptic feedback reflects only the contact forces from the environment, rather than the initial grasping force, the Tactile Mapping Node maps the change in the tactile force vector, $\Delta\mathbf{F}_{\text{sensor}}$, relative to a baseline. However, we observed that after forceful interactions, the tactile sensor's skin could undergo minor, non-recoverable deformations, causing its baseline to drift. 
To address this, we use the wrist-mounted F/T sensor to detect contact/non-contact states. Whenever the force magnitude remains below a predefined threshold (indicating the gripper is in free space), the tactile sensor's baseline is recalibrated to its current reading.

In the rendering process, we first transform $\Delta\mathbf{F}_{\text{sensor}}$ from the sensor's coordinate frame to the device's coordinate frame using a predefined rotation matrix $\mathbf{R}$:
$$ \mathbf{F}_{\text{device}} = \mathbf{R} \cdot \Delta\mathbf{F}_{\text{sensor}} $$
This 3D vector is then projected onto the device's 2D feedback plane to obtain a planar vector $\mathbf{f}_{\text{2D}} = [f_x, f_y]^T$. This 2D vector directly maps to the device's two outputs: its direction determines the servo motor's target angle $\theta$, and its magnitude determines the LRA's vibration amplitude $A$, scaled by a gain factor $k$.
$$ \theta = \text{atan2}(f_y, f_x) \quad \text{and} \quad A = k \cdot ||\mathbf{f}_{\text{2D}}|| $$

Because the hand-tracking teleoperation system maintains a constant relative orientation between the robot's end-effector and the operator's hand, the rotation matrix $\mathbf{R}$ is constant during teleoperation. This allows us to deliberately configure this transformation on a per-task basis to maximize feedback relevance. For the insertion tasks, for example, we define $\mathbf{R}$ such that the two most critical force directions—the primary insertion axis and the axis corresponding to the object's highest vulnerability to bending—are mapped directly onto the device's $x$-$y$ plane. This ensures the operator receives the most crucial information for the task.

\section{Experimental Evaluation}
\label{sec:evaluation}

We conducted a series of experiments to evaluate the \alg system from three perspectives: \textbf{(Experiment 1)} its core ability to convey directional information, \textbf{(Experiment 2)} its impact on human performance during contact-rich teleoperation tasks, and \textbf{(Experiment 3)} its potential to improve the quality of demonstrations for downstream imitation learning.
All human subject procedures were approved by the University of Chicago Institutional Review Board (IRB) under protocol IRB25-1498.
\subsection{Experiment 1: Directional Rendering Performance}
\label{subsec:exp1}

\textbf{Objective:} To quantitatively measure whether participants can understand directional cues rendered by the \alg device.

\textbf{Task:} Participants performed two direction-identification tasks: (1) a four-alternative forced-choice (4-AFC) task involving the four cardinal directions (up, down, left, right); and (2) a more challenging eight-alternative forced-choice (8-AFC) task that also included the four intercardinal directions. 

\textbf{Participants and Procedure:} Twelve participants were recruited for the study (nine male, three female; all aged 20--30). All reported having normal tactile sensation and no known sensorimotor impairments.
Each participant completed both the 4-AFC and 8-AFC tasks, with the order of the two tasks counterbalanced across participants to mitigate learning effects. The 4-AFC task consisted of 16 trials (4 directions $\times$ 4 repetitions), while the 8-AFC task also had 16 trials (8 directions $\times$ 2 repetitions). Within each task, the sequence of trials was fully randomized for each participant. During the experiment, participants were blindfolded to ensure reliance on haptic cues alone. At the beginning of each trial, we spun the rotor to several random positions before settling on the target direction to mask any auditory or proprioceptive cues. We then rendered a single directional vibration, which the participant identified from the available choices.

\textbf{Metrics:} We measured the overall accuracy for both tasks and computed confusion matrices to analyze error patterns across vibration directions.

\subsection{Experiment 2: Teleoperation Performance}
\label{subsec:exp2}
\textbf{Objective:} To evaluate the effectiveness of the \alg system in improving operator performance during contact-rich manipulation tasks.

\textbf{Tasks:} We designed three tasks to assess performance across different challenges in contact-rich manipulation.
\begin{itemize}
    \item \textbf{Key Insertion:} The task involves inserting a fragile wooden key into a 3D-printed lock with tight clearance. It requires precise alignment and controlled insertion forces to manage friction while avoiding excessive contact forces and bending torques that could damage/break the key. A trial was considered successful if the key was fully inserted without any visible damage.%
    \item \textbf{USB Insertion:} Participants inserted a standard USB-A plug into a desktop computer port. This everyday task requires accurately aligning the connector and applying sufficient force to overcome internal friction and retention features (e.g., side latches or screw caps added to standard USB ports to prevent accidental disconnection), while avoiding excessive force that could damage the plug or port. Success was defined by the plug being securely seated without damage. %
    \item \textbf{Spaghetti Probing:} Inspired by subsurface exploration and surgical interventions~\cite{gerovichev2002effect, chen2023force}, this task requires inserting a thin, fragile object into an opaque container of rigid objects with complex geometry.
    We employ uncooked spaghetti as the probe and use rice and building blocks to model granular media and rigid obstacles.
    Operators need to discern between the granular medium and rigid obstacles, adjusting their strategy to find a clear path to the bottom. A trial is deemed successful if the probe reaches the bottom without fracturing.
\end{itemize}

\textbf{Conditions:} We assessed expert performance under four feedback modalities: \textit{Vision-Only (C1)}, \textit{Vision + Non-directional Tactile (C2)}, \textit{Vision + Non-directional Controller Tactile (C3)}, and \textit{Vision + Directional Tactile (C4)}. The primary comparison is among (C1), (C2), and (C4), which are matched on all factors except for the type of haptic feedback provided. We additionally included (C3) as it represents a widely adopted baseline in teleoperation systems~\cite{kamijo2024learning, pattabiraman2024learning}.

\begin{itemize}
    \item \textbf{Vision-Only (C1):} The operator received only visual feedback. The \alg device was worn to control for weight and ergonomics, but remained inactive.
    \item \textbf{Vision + Non-directional Tactile (C2):} Our primary baseline. The operator wore the \alg device, which provided non-directional vibration with an amplitude proportional to the contact force magnitude. The rotor remained stationary.
    \item \textbf{Vision + Non-directional Controller Tactile (C3):} To compare against a common practice, operators held the Meta Quest controller for teleoperation. Haptic feedback was provided by the controller's built-in non-directional vibration motor. It is important to note that this condition introduces a change in the tracking modality from vision-based hand tracking (used in C1, C2, and C4) to more stable controller-based tracking.
    \item \textbf{Vision + Directional Tactile (C4, Proposed Method):} The operator used the full \alg system, receiving both magnitude and directional haptic feedback.
\end{itemize}

\textbf{Participants and Procedure:} For the \textit{Key Insertion} and \textit{USB Insertion} tasks, two expert participants, both with extensive experience using the teleoperation system, performed the experiments under all four feedback conditions. For data collection, we recorded 20 trials per condition for the \textit{Key Insertion} task and 60 trials per condition for the \textit{USB Insertion} task. Trials were evenly distributed between the two participants. To mitigate bias and learning effects, the experimental procedure was structured as follows: Trials were presented in blocks of five for a single condition. After each block, the feedback condition was changed, and the presentation order of these condition blocks was randomized for each participant. At the start of each trial, the robot gripper held the object securely. To introduce variation, we randomized the initial pose of the gripper within a small volume relative to the target (a $0.1\,\text{m} \times 0.1\,\text{m} \times 0.1\,\text{m}$ cube for USB Insertion and a $0.05\,\text{m} \times 0.05\,\text{m} \times 0.05\,\text{m}$ cube for Key Insertion). For the \textit{Spaghetti Probing} task, four non-expert participants completed the task under all four feedback conditions presented in randomized order. For each condition, participants performed ten probing trials, with the positions of rigid obstacles randomized across trials to prevent memorization effects.

\textbf{Metrics:} We measured task success rate, completion time, contact duration, maximum contact force, and maximum bending torque during the task. The maximum contact force is the maximum magnitude of the force vector, $\mathbf{F}$, read directly from the wrist-mounted F/T sensor. The maximum bending torque is computed from the F/T sensor readings $(\mathbf{F}, \boldsymbol{\tau})$ by first translating the torque to the gripper's contact point (offset by a vector $\mathbf{r}$) and then projecting it onto the object's  primary bending axis, $\hat{\mathbf{u}}$:
$$ m_{\text{bend}} = |\hat{\mathbf{u}} \cdot \boldsymbol{\tau}_{\text{grip}}| = |\hat{\mathbf{u}} \cdot (\boldsymbol{\tau} - \mathbf{r} \times \mathbf{F})|$$
Lower bending torque indicates safer, more aligned insertion.

\begin{figure*}[!t]
    \centering
    \includegraphics[width=\textwidth]{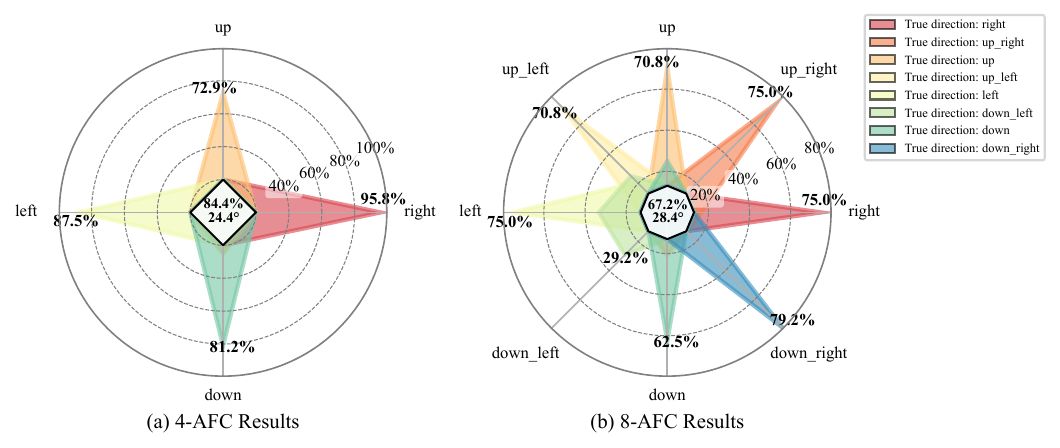}
    \caption{Direction-identification results presented as radar plots of the confusion matrices for the (a) 4-AFC and (b) 8-AFC tasks. Each colored wedge shows the distribution of participant responses for a given true direction. The overall accuracy and mean angular error across all trials are displayed in the center of each plot.}
    \vspace{-10pt}
    \label{fig:direction_results}
\end{figure*}

\subsection{Experiment 3: Imitation Learning Evaluation}
\label{subsec:exp3}
\textbf{Objective:} To evaluate whether directional tactile feedback during teleoperation improves the data quality for training imitation learning policies.

\textbf{Simplified Task Setup:} For this evaluation, we employed a constrained version of the \textit{Key Insertion} task to facilitate policy learning. The lock was rigidly fixed to the workspace, and the gripper's initial pose was pre-aligned with the keyhole. During teleoperation, the operators controlled the position of the gripper while its orientation was fixed. Despite these simplifications, the task remains non-trivial due to the tight clearance; safe insertion still requires reactive adjustments based on tactile feedback, as open-loop trajectories based solely on initial visual alignment frequently lead to excessive forces and key fracture.

\textbf{Procedure:} We trained two imitation learning policies using 10 demonstrations collected under conditions C3 and C4, respectively. The policies take tactile signal and robot proprioception as input, and predict an action chunk~\cite{zhao2023learning} of relative motion. The model architecture consists of separate MLP-based tactile and proprioception encoders, whose output is concatenated and fed to an MLP action decoder. We evaluate each policy over 10 trials.

\textbf{Metrics}: We measured rollout success rate, maximum contact force, and maximum bending torque for each policy.

\section{Results}
\label{sec:results}

\subsection{Directional Rendering Results}

The results of the directional rendering experiment (Fig.~\ref{fig:direction_results})  show that participants can interpret the directional cues from the \alg device with accuracies significantly above chance level in both the 4-AFC and 8-AFC tasks. 

In the 4-AFC task, participants achieved a high overall accuracy of $84.4\%$, with a mean angular error of $24.4^{\circ}$. As shown in Fig.~\ref{fig:direction_results}(a), performance was particularly strong for the left-right directions, with accuracies of $95.8\%$ for \texttt{right} and $87.5\%$ for \texttt{left}. In the more challenging 8-AFC task, which included intercardinal directions, the overall accuracy was $67.2\%$ with a mean angular error of $28.4^{\circ}$. While lower than the 4-AFC task, this result is substantially higher than the $12.5\%$ chance level, demonstrating that the device can effectively render eight distinct directions.

Error patterns reveal confusion between the \texttt{up} and \texttt{down} directions. We attribute this primarily to the mechanical limitation in the \alg device: the up-down axis is aligned with the driving belt's tension, which dampens the vibration and reduces the clarity of the directional feedback.

Second, in the 8-AFC task, the accuracy for the \texttt{down-left} direction was notably lower at $29.2\%$, with it most often being mistaken for \texttt{left}. This is likely a compound effect. With the already attenuated vertical component, the stimulus for \texttt{down-left} becomes dominated by its horizontal component. For a right-handed grasp, the perceived directional cue associated with \texttt{down-left} may induce a torque on the arm that is perceptually similar to that of a pure \texttt{left} cue, potentially leading to this confusion.\todomw{This description could be simplified} %

\subsection{Teleoperation Results}
\begin{table*}[htbp]
    \centering
    \caption{Teleoperation Performance on All Tasks}
    \label{tab:teleop_results}
    \setlength{\tabcolsep}{2.5pt}
    \begin{tabular*}{\textwidth}{clcccc}%
        \toprule
        Task & Condition & Success Rate $\big\uparrow$ & Completion Time (s) $\big\downarrow$ & Contact Duration (s) $\big\downarrow$ & Max Force (N) $\big\downarrow$ \\
        \midrule
        \multirow{4}{*}{Key Insertion} & Vision-Only (C1) & 63.2\% (12/20) & 12.92 $\pm$ 5.56 & 8.80 $\pm$ 4.48 & 97.6 $\pm$ 89.7 \\
        & Vision + Non-directional Tactile (C2) & 73.7\% (14/20) & 13.54 $\pm$ 6.52 & 8.50 $\pm$ 5.82 & 104.9 $\pm$ 67.5 \\
        & Vision + Non-directional Controller Tactile (C3) & 90.0\% (18/20) & \textbf{10.39 $\pm$ 3.08} & \textbf{6.25 $\pm$ 3.16} & 79.6 $\pm$ 71.9 \\
        & \textbf{Vision + Directional Tactile (C4)} & \textbf{100.0\% (20/20)} & 11.61 $\pm$ 3.05 & 7.00 $\pm$ 3.61 & \textbf{65.9 $\pm$ 59.6} \\
        \midrule
        \multirow{4}{*}{USB Insertion} & Vision-Only (C1) & 93.3\% (56/60) & 14.39 $\pm$ 4.50 & 7.01 $\pm$ 5.26 & 89.8 $\pm$ 29.4 \\
        & Vision + Non-directional Tactile (C2) & 86.7\% (52/60) & 15.45 $\pm$ 6.94 & 7.67 $\pm$ 6.87 & 81.9 $\pm$ 32.3 \\
        & Vision + Non-directional Controller Tactile (C3) & \textbf{98.3\% (59/60)} & \textbf{11.11 $\pm$ 3.81} & \textbf{3.95 $\pm$ 3.76} & 82.9 $\pm$ 27.6 \\
        & \textbf{Vision + Directional Tactile (C4)} & \textbf{98.3\% (59/60)} & 12.84 $\pm$ 3.45 & 5.54 $\pm$ 3.47 & \textbf{79.8 $\pm$ 23.2} \\
        \midrule
        \multirow{4}{*}{Spaghetti Probing} & Vision-Only (C1) & 57.5\% (23/40) & 22.57 $\pm$ 10.37 & 14.27 $\pm$ 11.94 & 10.9 $\pm$ 5.2 \\
        & Vision + Non-directional Tactile (C2) & 55.0\% (22/40) & 25.72 $\pm$ 10.35 & 17.76 $\pm$ 11.54 & 12.2 $\pm$ 6.2 \\
        & Vision + Non-directional Controller Tactile (C3) & 42.5\% (17/40) & \textbf{20.63 $\pm$ 10.15} & \textbf{12.75 $\pm$ 9.69} & \textbf{9.2 $\pm$ 8.4} \\
        & \textbf{Vision + Directional Tactile (C4)} & \textbf{60.0\% (24/40)} & 24.36 $\pm$ 12.33 & 16.45 $\pm$ 13.21 & 10.6 $\pm$ 5.5 \\
        \bottomrule
    \end{tabular*}
\end{table*}

\begin{figure}[!t]
    \centering
    \includegraphics[width=\columnwidth]{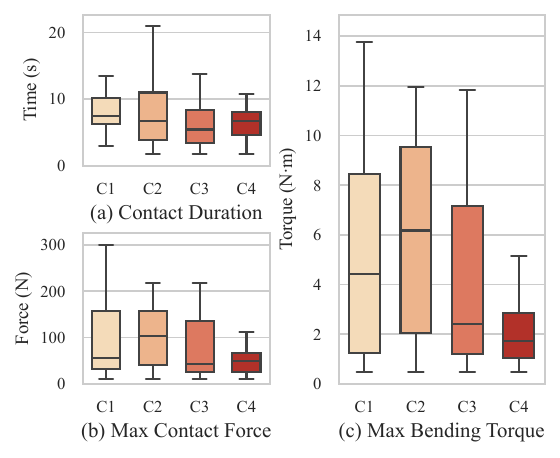}
    \caption{Teleoperation performance on the \textit{Key Insertion} task across four teleoperation conditions (C1-C4). (a) Contact duration distributions. (b) Maximum contact force distributions. (c) Maximum bending torque distributions.}
    \label{fig:key_insertion_boxes}
\end{figure}

Table~\ref{tab:teleop_results} summarizes the quantitative results of the teleoperation experiment. We observe the highest success rates in the C4 condition across all tasks, most notably in \textit{Key Insertion} (100\% vs. 63.2\% for C1).
A similar trend was observed in the \textit{USB Insertion} task, where both our method (C4) and the controller-based method (C3) resulted in near-perfect success rates (98.3\%).
Regarding efficiency, while C3 yielded the shortest completion and contact times, this is likely influenced by the more stable tracking modality (e.g., IMUs) used exclusively in C3, as noted in Section~\ref{subsec:exp2}; our method (C4) exhibited the highest efficiency among the vision-based hand-tracking conditions (C1, C2, and C4).

Further performance differences were observed in the operators' ability to perform fine force control.
As summarized in Table~\ref{tab:teleop_results}, C4 was associated with the lowest maximum contact force in both the \textit{Key Insertion} and \textit{USB Insertion} tasks.
The box plots in Fig.~\ref{fig:key_insertion_boxes} also show that C4 produced lower bending torque in the \textit{Key Insertion} task, which is an important indicator of insertion quality and safety. 
These observations suggest that, for our expert users, directional feedback may have contributed to gentler and more consistent manipulations, reducing the risk of damaging fragile components.

In the \textit{Spaghetti Probing} task, C4 reached the highest success rate (60.0\%).
Although C3 recorded lower times and forces in its successful trials, this likely stems from selection bias; given its low success rate (42.5\%), C3's metrics likely originate from a small number of trials where the operator found an obstacle-free path on their first insertion attempt, thus avoiding any time-consuming re-probing that occurs after hitting an obstacle. In contrast, C4's higher success rate suggests that directional feedback allows operators to detect and navigate around obstacles more effectively.

\subsection{Imitation Learning Results}

\begin{table}[]
    \centering
    \caption{Imitation Learning Results on Key Insertion Task}
    \label{tab:learning_results}
    \setlength{\tabcolsep}{2.5pt}
    \begin{tabular}{lccc}
        \toprule
         Condition & Success Rate$\big\uparrow$ & Max Force(N)$\big\downarrow$ & Max Bend.\ Torque (N$\cdot$m)$\big\downarrow$ \\
         \midrule
         C3 & 60\% & 65.81$\pm$33.70 & 5.26$\pm$3.15 \\
         \textbf{C4} & \textbf{90\%} & \textbf{24.62$\pm$14.99} &  \textbf{3.79$\pm$3.00}\\         
         \bottomrule
    \end{tabular}
\end{table}

Table~\ref{tab:learning_results} summarizes the learning performance on the simplified \textit{Key Insertion} task. The policy trained on C4 demonstrations achieved a 90\% success rate, outperforming the C3 baseline (60\%) while recording lower peak forces and torques. Qualitatively, we observed the C4 policy adjusting the gripper's position in response to sensed contact forces. In contrast, the C3 policy lacked this reactive behavior, often leading to excessive contact forces and insertion failure.

\section{Conclusion}
\label{sec:conclusion}
We introduced \alg, a novel, low-cost wearable haptic device that delivers 2D directional cues by mechanically rotating a single LRA. We presented the full design and integration of \alg and evaluated its effectiveness through three studies. A perception experiment confirmed that users can reliably perceive directional cues, while teleoperation experiments demonstrated that \alg improves success rates and reduces completion times and contact forces in contact-rich tasks. Finally, preliminary evaluations suggest that demonstrations collected with \alg lead to imitation learning policies with higher success rates and lower contact intensity.

Our study has three main limitations. First, the current hardware is restricted to rendering directional cues within a 2D plane rather than 3D spatial directions. Second, despite system optimizations, some end-to-end latency remains, and its effect on teleoperation performance is not characterized. Third, the evaluation was limited to three teleoperation tasks and one imitation learning task, leaving questions about generalizability to more diverse manipulation scenarios.

As for future work, we plan to explore several directions:
\begin{itemize}
    \item \textbf{More feedback modalities:} Extend \alg by integrating additional actuators for normal-force cues.
    \item \textbf{Impact on learning:} Conduct a more systematic investigation into how haptic feedback influences the quality of demonstration datasets and the resulting performance of learned policies.
\end{itemize}

\section*{ACKNOWLEDGMENT}
We would like to extend our gratitude to Yudai Tanaka, Shan-Yuan Teng and Romain Nith of the Human Computer Integration Lab for their feedback and assistance. %

{\small
\balance
\bibliographystyle{IEEEtranN}
\bibliography{refs}
}

\end{document}